\documentclass{article}
\usepackage{graphicx} 
\usepackage{setspace}
\usepackage{amsmath}
\usepackage{amssymb}
\usepackage{siunitx}
\usepackage{biblatex}
\usepackage{comment}
\usepackage{geometry}
\usepackage{hyperref}
\usepackage{lineno}
\usepackage{authblk}
\title{Learning a Stochastic Differential Equation Model of Tropical Cyclone Intensification from Reanalysis and Observational Data}
\author[1]{Kenneth Gee}
\author[1]{ Sai Ravela}
\affil[1]{Earth Signals and Systems Group, Massachusetts Institute of Technology,  Cambridge, MA 02139, USA}

\usepackage{setspace} 

\AtBeginBibliography{\singlespacing}

\begin{document}

\doublespacing
\maketitle

\begin{abstract}
    Tropical cyclones are among the most consequential weather hazards, yet estimates of their risk are limited by the relatively short historical record. To extend these records, researchers often generate large ensembles of synthetic storms using simplified models of cyclone intensification. Developing such models, however, has traditionally required substantial theoretical effort. Here we explore whether equation-discovery methods, a class of data-driven techniques designed to infer governing equations, can accelerate the process of developing simplified intensification models. Using observational storm data (IBTrACS) together with environmental conditions from reanalysis (ERA5), we learn a compact stochastic differential equation describing tropical cyclone intensity evolution. We focus on TCs because their dynamics are well studied and a hierarchy of reduced-order models exist, enabling direct comparison of the learned model to physically-derived counterparts. We find that the learned model simulates synthetic TCs whose intensification statistics and hazard estimates are consistent with observations and competitive with a leading physics-based TC intensification model. Our model also reproduces known nonlinear dynamical behavior of tropical cyclones, including as a saddle node bifurcation as inner core ventilation is increased. This result shows that equation-discovery approaches, when applied directly to storm intensity, can recover not only realistic statistics but also physically meaningful dynamical structure. These findings highlight the potential for data-driven methods to complement existing theory and reduced-order models in the study of extreme weather.
\end{abstract}

\section{Introduction}

Tropical Cyclones (TCs) are among the costliest and deadliest natural disasters \cite{needham_review_2015, qiu_decades_2025}. To plan for and mitigate the risk posed to coastal communities by TCs, it is important to estimate their intensity at landfall. Storm return periods and intensity distributions at landfall are common metrics of hazard, but the return periods of the most intense storms exceed the duration of high quality intensity observations ($\sim40$ years \cite{knapp_international_2010}) which limit the strength of such estimates from historical data alone \cite{peduzzi_global_2012}.

One way to fill this data gap is to use models to simulate a large collection of realistic TC tracks and intensities. Tropical cyclones emerge in Global Circulation Models, so one might be tempted to run GCMs for a long time and extract storm data. However, this approach is computationally infeasible as GCMs at the resolution to resolve TCs are prohibitively expensive \cite{acosta_computational_2024}. Sacrificing resolution for computational speed also fails because the dynamics of the inner eyewall are crucial for intensification \cite{willoughby_concentric_1982} but requires kilometer-scale resolution - the Integrated Forecasting System (IFS) run at $1/4\si{\degree}$ resolution \cite{ecmwf_ifs_2024} systemically underestimates intensity for this reason \cite{hodges_how_2017}. However, intensification is driven in large part by the large-scale environment surrounding a storm \cite{emanuel_environmental_2004}, enabling the development of inexpensive models of intensity which depend only on coarse-grained features of the environment \cite{emanuel_climate_2006} which still accurately capture the statistics of intensity. Coupled with separate track generation and cyclogenesis models, ``downscaling" models of tropical cyclones have been developed \cite{lin_open-source_2023,lee_environmentally_2018, bloemendaal_generation_2020} which generate completely synthetic storms. They have been used to generate a dataset of 10,000 years of storms \cite{bloemendaal_generation_2020} in present climate for TC hazard estimation, and to generate TC hazard estimates in future climate scenarios where no data is available \cite{lin_response_2025}.

The intensity, track and genesis components have all developed appreciably over recent decades, but intensification remains the most challenging because it depends on several nonlinear and challenging to parameterize processes \cite{emanuel_predictability_2016}. The intensity, here measured as the maximum sustained azimuthal wind speed, is nonlinear because it depends on a positive feedback between surface wind speed and enthalpy extraction from the surface \cite{emanuel_thermodynamic_1999}, and the parameterization of this exchange depends on challenging-to-constrain details of sea spray in high winds \cite{emanuel_tropical_2003}. Further, the impact of synoptic vertical wind shear on intensity depends on parameterizations of turbulent downdrafting of low-entropy environmental air into the boundary layer \cite{tang_ventilation_2012}.

What, then, is the best way to model intensification? Many approaches have been developed to answer this question. Several authors have trained autoregressive models of intensity, which are linear models that forecast intensity at discrete increments of time as a function of lags of intensity and large-scale environmental variables \cite{lee_autoregressive_2016,bloemendaal_generation_2020,demaria_statistical_1994}. Such models often match the statistics of observed tracks closely but can generalize poorly in regions with little training data \cite{meiler_intercomparison_2022}. StormCast is a machine learning emulator of a Convection Allowing Model (CAM) of atmospheric dynamics which can rapidly generate detailed cyclone dynamics (including genesis, track and intensity at once) \cite{pathak_fourcastnet_2022}, but such models may also struggle to generalize to out-of-training data atmospheric states such as those from future climate scenarios, and StormCast doesn't couple to the underlying ocean which is a key driver of intensity \cite{schade_oceans_1999}. Finally, physical models have been developed, such as the Coupled Hurricane Intensity Prediction System (CHIPS) \cite{emanuel_environmental_2004} which couples an intensity model to a simplified 1d model of the upper ocean, and FAST \cite{emanuel_fast_2017} which is an Ordinary Differential Equation model arrived at empirically to inexpensively emulate CHIPS. Both models were developed using physical understanding and hence are expected to generalize well. They can also be used to deepen understanding of TC intensification dynamics through a nonlinear dynamics analysis \cite{slyman_tipping_2023}, but exhibit biases such as underestimation of extreme intensities \cite{lin_open-source_2023}.

An approach which has not been attempted is to learn a physics-style model, such as an ordinary differential equation (ODE) or stochastic differential equation (SDE), from observational cyclone intensity and reanalysis data directly. Such a model would be amenable to mathematical analysis and would fit well to data by construction. A wealth of literature has developed methodologies to learn such models. The Nonlinear Autoregressive Moving Average model with Exogenous inputs (NARMAX) paradigm \cite{noauthor_identification_2013} refers to a large body of techniques for learning nonlinear autoregressive and nonlinear continuous models using time series data. More recently, interest has been rekindled with the Sparse Identification of Nonlinear Dynamical systems (SINDy) framework \cite{brunton_discovering_2016} which regresses finite-difference quotient estimates of the system's derivative onto a large library of candidate features. Model structure is learned by imposing a sparsity constraint on regression so that only a small subset of the initial feature library are ultimately utilized. SINDy, however, is not well-suited to the intensification problem because observations of TC intensity are discretized to $5$ knot bins and available at coarse temporal resolution of $6\si{\hour}$, making finite-difference quotient intensity estimates very noisy. Integral SINDy instead uses integrations of the underlying differential equation to construct a sparse linear regression problem which is more robust to observation noise \cite{schaeffer_sparse_2017, messenger_weak_2021}. However, this approach suffers from performing quadrature at the resolution of observations which compromises integration quality in temporally coarse settings. The Informative Ensemble Kalman Learner (IEKL) is another approach to system identification which instead integrates ODEs at arbitrary numerical precision and uses observations whenever they are available to perform parameter estimation and structure learning \cite{trautner_informative_2020}. Given its numerical flexibility, the IEKL has the potential to recover dynamical systems in a wider range of cases than previous approaches.

In this paper, we show that nonlinear system identification techniques can learn a physics-style model of intensity from observational data and simulations. We present a 10-term polynomial stochastic differential equation of intensity which depends on estimates of the large scale environment surrounding the storm. The model was learned directly from a synthesis of high-resolution simulations and observations, specifically the reanalysis climate model ECMWF Reanalysis 5 (ERA5) \cite{hersbach_era5_2023} and the observational dataset of hurricane intensity International Best Track Archive for Climate Stewardship (IBTrACS) \cite{knapp_international_2010}. Methodologically, we solve an Integral SINDy problem then fine tune parameters using the inference stage of the IEKL. The model produces intensity time series which match many aspects of the statistics of observed storms. The model is interpretable in the context of existing theoretical understanding of TC intensification and produces reasonable estimates of TC return periods and landfalling intensity distribution. While these results depend on pre-engineered features, they demonstrate promise that such system identification methods can accelerate the scientific process of identifying low order models of complex earth system dynamics.

In Section \ref{sec:intensification-features}, we review the environmental variables which drive the intensification model and introduce the equation structure. In Section \ref{sec:learned-model}, we present the learned equation, analyze its bifurcations, and compare the climatology and individual intensification time series with those from the historical record. In Section \ref{sec:equation-learning}, we detail the model training procedure and sparsity promotion step. In Section \ref{sec:discussion}, we discuss our results and point to future directions.

\section{Intensification Features}
\label{sec:intensification-features}

The model depends on a set of environmental features that are known to drive TC intensification. These features are already ``engineered" based on previous scientific effort and are not inferred from primitive variables. Each feature is estimated from ERA5 reanalysis and forces intensification along a track.

The first feature is the Potential Intensity $V_p$, a maximum possible intensity given the vertical temperature profile surrounding the storm \cite{bister_dissipative_1998}. A Tropical Cyclone converts thermal energy from the warm sea surface to kinetic energy by releasing the heat into the cool upper atmosphere. A thermodynamic argument constrains the maximum possible kinetic energy using the temperature profile of the air column. A cyclone is often thought of as approaching its potential intensity, but its ability to achieve this maximum is bounded by other processes.

The second feature captures the impact of the upper ocean on TC intensity. Tropical cyclone winds stir up the underlying ocean and upwell cold waters from below the mixed layer. This cools the sea surface and cuts off the storm's energy source, damping the cyclone. This effect is strongest with a shallow mixed layer $h_m$, steep sub-mixed layer stratification $\Gamma$, and a slow cyclone velocity (not wind speed) $u_T$, and strong intensity $v$.

An estimate of the climatological upper ocean heat content as a function of the track and environment is given by

\begin{equation}
    z=0.01\Gamma^{-0.4}h_m u_TV_p^{-1}
\end{equation}

which takes positive values and is treated as dimensionless \cite{schade_oceans_1999}. The impact of intensity $v$ is incorporated in another nondimensional parameter 

\begin{equation}
    \alpha=1-0.87\exp(-z/v)
\end{equation}

which is bounded to $\alpha\in[0.13,1]$ and where zero damping corresponds to unity \cite{schade_oceans_1999}. Only $\alpha$ is used directly in the model, $z$ is an internal variable in the computation of $\alpha$.

The third feature is the environmental mid-level entropy deficit $\chi$, which measures the degree to which the air surrounding the cyclone is cooler and drier than that in its inner core. More specifically, it measures the difference in entropy at $600\si{\hecto\pascal}$ between the cyclone's inner core and surrounding environment \cite{tang_sensitivity_2012}. High values indicate that intruding air from the environment will dry out the inner core which limits moist convection and damps intensity.

The final physical variable is synoptic vertical wind shear $S$. Synoptic wind shear is thought to tilt the moist inner core of the cyclone which produces downdrafts that advect aforementioned lower entropy environmental air into the boundary layer where that air is taken up by the cyclone's core \cite{tang_ventilation_2012}, decreasing intensity by the mechanism explained above. Shear is estimated from 12 hourly ERA5 snapshots, and during integration the snapshot nearest in time is utilized. The velocity is averaged at the $250\si{\milli\bar}$ and $850\si{\milli\bar}$ pressure levels in a $6^\circ$ radius about the center of the storm, following \cite{chen_effects_2006}, and shear is computed as $S=|\vec{u}_{250}-\vec{u}_{850}|$. All other features are computed using monthly averaged ERA5 variables.

We use the FAST open source codebase \cite{lin_open-source_2023} to compute these features and refer the reader to this source for further implementation details.

\subsection{Equation Setup}

Tropical cyclone intensity evolves dynamically as a function of the cyclone's internal state as well as the large-scale environment through which it moves. Following \cite{lin_open-source_2023}, we assume that a differential equation model of intensity $v$

\begin{equation}
    \dot{v}=F(v,\alpha,V_p,\chi,S)
\end{equation}

can capture observed dynamics, where $F$ is some function of the current intensity and aforementioned environmental variables. We nondimensionalize each variable by dividing by a characteristic scale 

\begin{equation}
    v'=\frac{v}{50\si{\meter\per\second}}, V_p'=\frac{V_p}{50\si{\meter\per\second}},\chi'=\frac{\chi}{2\si{\joule\per\kelvin\per\kilo\gram}},S'=\frac{S}{10\si{\meter\per\second}}
\end{equation}

so that each variable is $O(1)$. We parameterize the function $F$ as a polynomial of the input variables up to a particular degree. The functional relationship is 

\begin{equation}\label{eq:model_setup}
    dv'=(\underbrace{\beta_0+\beta_1v'+\beta_2v'^2+\beta_3v'V_p' + \beta_4v'V_p'^2+\cdots+\beta_{55}S'^3}_{56\text{ terms}})d\tau+\sigma'(v')dW_t
\end{equation}

where each feature is a product of the nondimensionalized input variables and $\tau=6\si{\hour}$ is the observational timescale. The term $\sigma'(v')$ is the standard deviation (of nondimensionalized intensity $v'$) of a Wiener Process learned during a calibration step to model remaining uncertainties in intensification, and are detailed in Sec. \ref{sec:calibration}.

In general, each feature is $\phi_j=v'^{k_1}\alpha^{k_2}V_p'^{k_3}\chi'^{k_4}S'^{k_5}$ for integers $\sum_{j}k_j\leq k_{max}=3$. All coefficients $\beta_i$ have units $\si{\tau^{-1}}$ by dimensional consistency. A range of polynomial degrees was attempted but validation error did not significantly decrease beyond degree 3.

Polynomial features are highly multicolinear so only a subset of terms may be necessary to capture the physical relationship. Hence, we seek a sparse solution, where a small subset of terms above have nonzero coefficients. Sparsity promotion is a means to regularize solutions and encourage generalization \cite{kutz_parsimony_2022}. We detail the sparsity promotion procedure in Sec. \ref{sec:equation-learning}.

\section{Learned Intensification Model}
\label{sec:learned-model}

Before we detail the methodology to learn the equation, we report it here. The model is 

\begin{equation}\label{eq:learned_sde}
    dv'= (\underbrace{\beta_0 v'\alpha V_p' + \beta_1v' + \beta_2 v'\alpha^2 + \beta_3 v'^3 + \beta_4\alpha V_p' + \beta_5\alpha V_p'^2 + \beta_6 V_p'^3 + \beta_7 V_p'^2S' + \beta_8\chi'^2 + \beta_9 \chi'^3}_\text{10 terms})d\tau+\sigma'(v')dW_t
\end{equation}

\begin{table}[]
    \centering
    \begin{tabular}{c | l}
        $v'\alpha V_p'$ & $\beta_0=1.53\times10^{-1} \si{\tau^{-1}}$ \\
        $v'$ & $\beta_1=-7.06\times10^{-2}\si{\tau^{-1}}$ \\
        $v'\alpha^2$ & $\beta_2=-6.70\times10^{-2}\si{\tau^{-1}}$ \\
        $v'^3$ & $\beta_3=-5.94\times10^{-2}\si{\tau^{-1}}$ \\
        $\alpha V_p'$ & $\beta_4=5.02\times10^{-2}\si{\tau^{-1}}$ \\ 
        $\alpha V_p'^2$ & $\beta_5=-4.70\times10^{-2}\si{\tau^{-1}}$ \\
        $V_p'^3$ & $\beta_6=1.26\times10^{-2}\si{\tau^{-1}}$\\
        $V_p'^2S$ & $\beta_7=-1.16\times10^{-2}\si{\tau^{-1}}$\\
        $\chi'^2$ & $\beta_8=-7.40\times10^{-3}\si{\tau^{-1}}$\\
        $\chi'^3$ & $\beta_9=1.20\times10^{-3}\si{\tau^{-1}}$\end{tabular}
    \caption{Coefficient values of the sparse, learned equation.}
    \label{tab:coefficient_value_table}
\end{table}

and includes 10 terms from the original 56. As it is defined in the continuum, it can generate intensity values at arbitrary temporal resolution.

\begin{figure}
    \centering
    \includegraphics[width=0.8\linewidth]{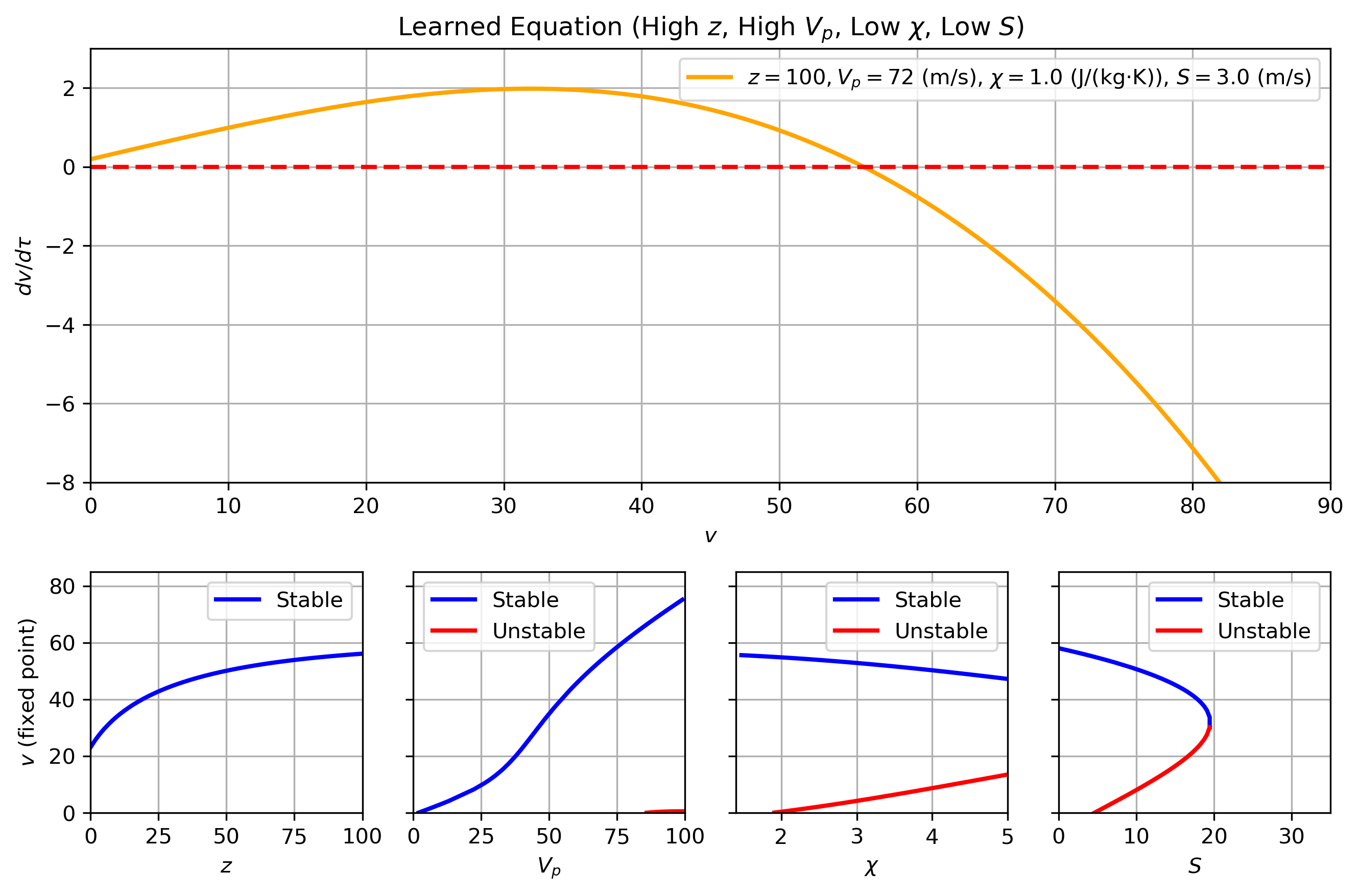}
    \caption{(Top) Plot of the learned differential equation as a function of intensity in conditions favorable to intensification, namely high upper ocean heat content $z=100.0$, high potential intensity $V_p=72\si{\meter\per\second}$, low environmental entropy deficit $\chi=1.0\si{\joule\per\kilo\gram\per\kelvin}$ and low wind shear $S=3.0\si{\meter\per\second}$. 
    (Bottom) Bifurcation diagrams of the model as all parameters but one are varied. Parameter ranges are the complete range the parameters experience over all tracks.}
    \label{fig:eq-bifurcation}
\end{figure}

The deterministic component of the equation characterizes a stable fixed point in intensity. In the top panel of Fig. \ref{fig:eq-bifurcation}, we plot the ODE as a function of intensity $v$ for a storm in conditions ripe for intensification, with high potential intensity $V_p$, high ocean heat content $z$, low environmental entropy deficit $\chi$ and low vertical wind shear $S$. A single, stable fixed point is present at $v=64.5\si{\meter\per\second}$ which the cyclone will steadily approach.

The stable fixed point varies as a function of the environmental parameters and may be annihilated in conditions unfavorable to intensification. In the bottom of Fig. \ref{fig:eq-bifurcation}, we plot changes in the fixed point as a single environmental parameter is varied through a series of bifurcation plots. The stable fixed point increases with increased ocean heat content $z$, increasing potential intensity $V_p$, decreasing entropy deficit $\chi$, and decreasing environmental wind shear $S$, in agreement with expectations. A separate unstable fixed point emerges and increases with increasing entropy deficit $\chi$ and wind shear $S$. Above a threshold wind shear $S=19\si{\meter\per\second}$ the lower unstable and upper stable fixed points annihilate in a saddle-node bifurcation. This bifurcation is a well-established aspect of TC nonlinear dynamics. It is predicted theoretically by modifying a TC's energy budget to include a sink term from mid-level downdrafts \cite{tang_midlevel_2010}, and is reproduced in numerical models \cite{tang_sensitivity_2012} and observational data \cite{tang_ventilation_2012}. The bifurcation is also present in the FAST model \cite{emanuel_physics_2026}. Existing theory parameterizes the bifurcation using the ventilation index \cite{tang_ventilation_2012}

\begin{equation}
    \Lambda=\frac{\chi S}{V_p}
\end{equation}

where higher values lead to the onset of the saddle-node bifurcation and the loss of stability of solutions. Increases in either $S$ or $\chi$ induce the bifurcation, a property reproduced in our model because for sufficiently high background $S$ values (we find $S>15\si{\meter\per\second}$), increases to the entropy deficit also induce the bifurcation.

\subsection{Intensification Dynamics Comparison}\label{sec:intensification-dynamics-comparison}

Here we compare the intensification of historical storms to what the learned model predicts. We pick Hurricane Katrina (2008), Hurricane Luis (1995), and Typhoon Haiyan (2013) as test cases because each is a historically significant and damaging storm. The chosen storms are not in the training or validation sets. In Fig. \ref{fig:track-comparisons} we plot the historical intensification and the probability contours of 100 synthetic storms generated along the same track using ERA5 reanalysis estimates of the surrounding environment. Each storm remains within the $2\sigma$ contour over $93.75\%$, $97.8\%$ and $100\%$ of their duration, respectively.

Both Hurricane Katrina and Typhoon Haiyan underwent Rapid Intensification - a process where a storm increases in intensity by at least $18\si{\meter\per\second}$ within a 24 hour period. The occurrence of Rapid Intensification empirically distinguishes the strongest storms from the weakest storms in terms of Lifetime Maximum Intensity (LMI) \cite{lee_rapid_2016}, hence its representation is key to capturing extreme events. While Rapid Intensification only occurs in large scale environmental conditions favorable to intensification \cite{kaplan_evaluating_2015}, as measured by the environmental features used in our model, its timing is very challenging to predict \cite{yang_long_2020} and, in high resolution numerical models, depends on state perturbations on the order of numerical modeling error \cite{zhang_effects_2013}. Hence, while our model can capture the large scale climatological dependence of RI, it is not expected to precisely predict RI timing. That rapid intensification is included within the $2\sigma$ ensemble suggests that the stochastic parameterization detailed in Sec. \ref{sec:calibration} can represent these challenging dynamics.

\begin{figure}
    \centering
    \includegraphics[width=1.0\linewidth]{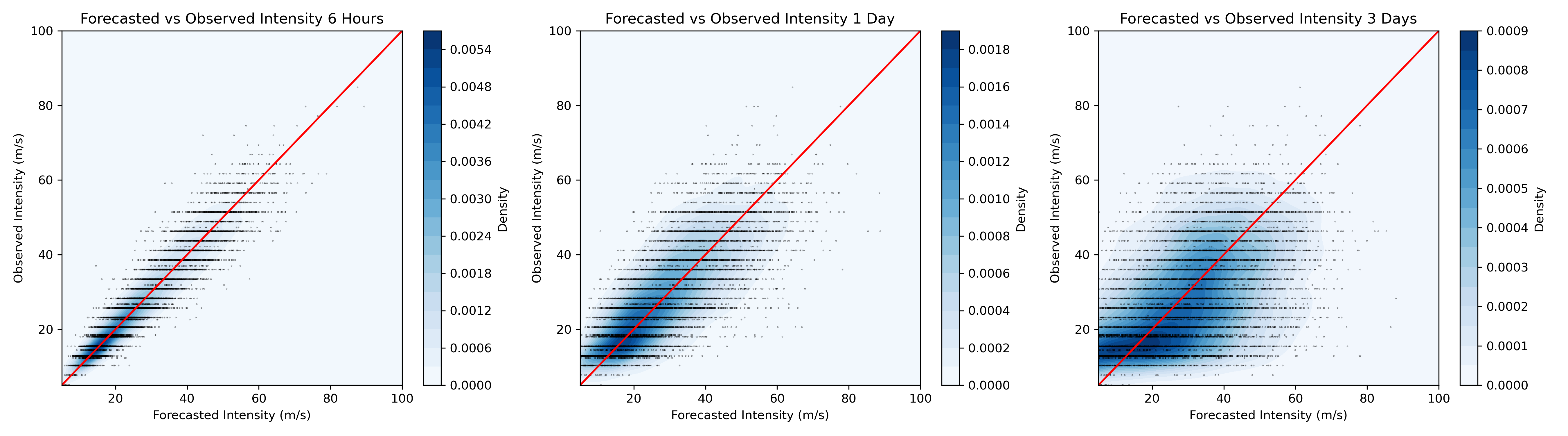}
    \caption{Scatter plot of forecasted versus observed IBTrACS intensities at time horizons of 6 hours, 1 day and 3 days on the testing dataset. A KDE estimate of the distribution is plotted in the background. IBTrACS is discretized to $5$ knot increments.}
    \label{fig:observed-true-scatter}
\end{figure}

Next, we more systematically assess if simulated intensification matches the distribution of historical intensification. On a testing dataset of all Northern Hemisphere storms from 1979 to 2015, we simulate intensity along a historical IBTrACS track at each initial condition along the track and then scatter plot pairs of the synthetic and observed intensity in Fig. \ref{fig:observed-true-scatter}. If the statistics of simulated storms match that of observed storms, then the distribution will be symmetric about the $x=y$ line. We report the plot at time horizons of $6\si{\hour}$, $24\si{\hour}$ and $72\si{\hour}$ and quantify the degree of symmetry by computing the Total Variation distance between a Gaussian KDE estimate of the distribution and its reflection. The Total Variation distance is

\begin{equation}
    d_{tv}(P,Q)=\frac{1}{2}\sup_x |P(x)-Q(x)|
\end{equation}

and measures the greatest discrepancy in probability afforded to the same event between two distributions. It is is zero if the distributions match perfectly and 1 if they have disjoint supports. The total variation distance is $d_{TV}=0.0529,0.0893,0.2137$ for each time horizon, respectively. A persistence model at $6\si{\hour}, 24\si{\hour}$ and $66\si{\hour}$ time horizons achieve $d_{TV}= 0.1912,0.2469,0.0813$, respectively. This shows that the skill of persistence models is weak at sub $1$ day time horizons, but that 3 day forecasts begin to decorrelate with initial intensities so their comparison approaches the distance between the marginal distribution of intensity with itself, which is $d_{TV}=0$. At $6\si{\hour}$ and $24\si{\hour}$, the distribution is strongly symmetrical, suggesting that the statistics of intensification match observations. At $72\si{\hour}$, the bulk of the distribution is symmetrical except for a significant overestimation of Tropical Depressions (TDs,$<18\si{\meter\per\second}$). However, hurricane agencies often don't report Tropical Depressions or do so inconsistently \cite{hodges_how_2017}, so our overestimation may reflect that if a storm becomes a Tropical Depression it is less likely to be reported in IBTrACS while our model's intensity is not lower bounded.

\begin{figure}
    \centering
    \includegraphics[width=1.0\linewidth]{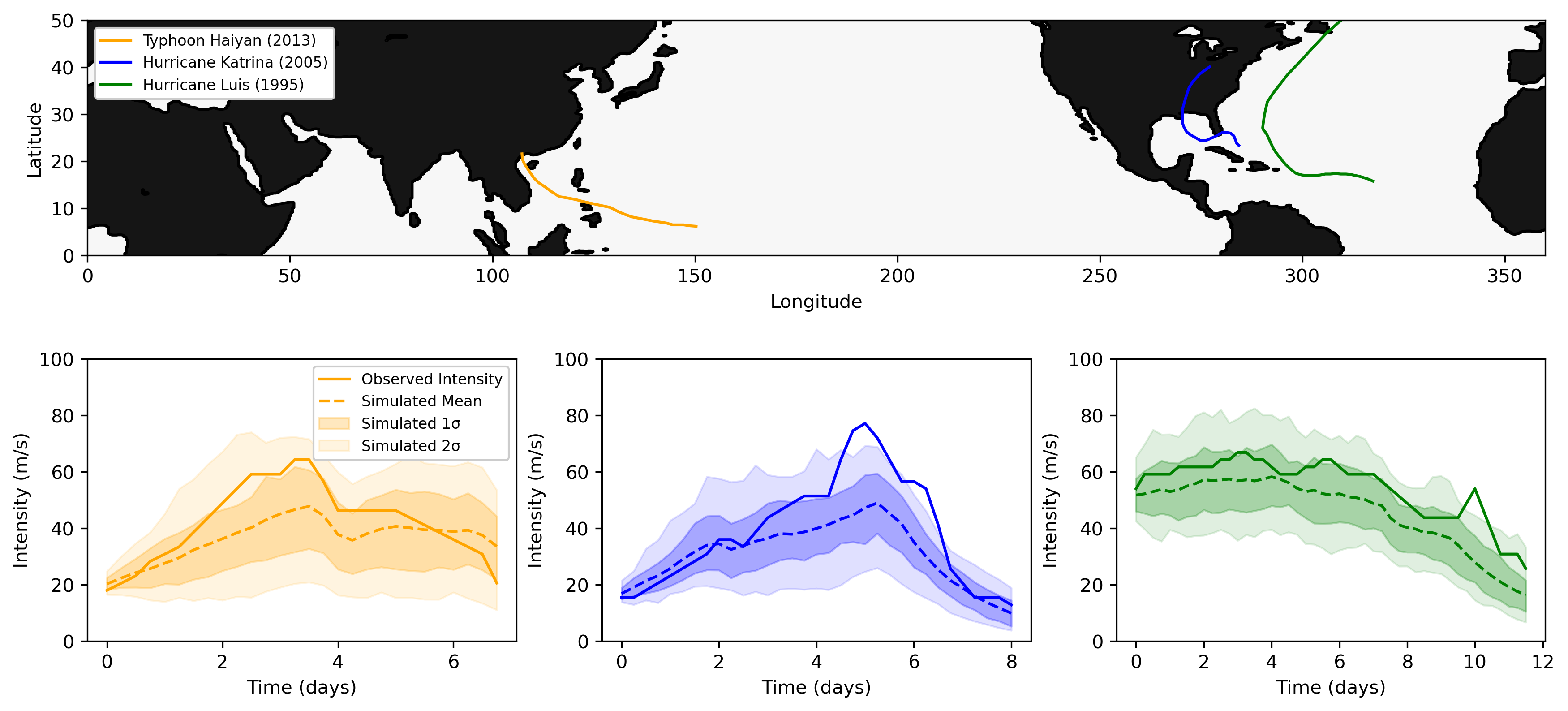}
    \caption{(Top) Three example Tropical Cyclone tracks from the IBTrACS dataset. (Bottom) IBTrACS intensities compared against the distribution of an ensemble of 100 synthetic storms for the same track and environmental forcings with confidence intervals.}
\end{figure}\label{fig:track-comparisons}

\subsection{Climatological Comparison}

Here, we assess how well the learned model represents TC hazard and climatology in comparison to observations and FAST. We generate 100 synthetic intensity time series per track over all IBTrACS tracks from 1979 to 2015 using the learned model. For the purposes of computing TC climatology, we treat this collection as 3600 years of storms (100 instances of the 36 year subset of IBTrACS). The intensity initial condition comes from the observed IBTrACS intensity (model stochasticity will wipe out this information before landfall), and environmental forcing comes from ERA5. We generate 820 years worth of synthetic storms using the Open-Source FAST codebase over the same set of ERA5 forcings \cite{lin_open-source_2023}. Note that our model simulates intensity along out-of-sample IBTrACS tracks, while the FAST pipeline generates tracks using separate track and cyclogenesis models. This isolates the skill of the intensification model by ensuring track and cyclogenesis statistics match observations, but gives a disadvantage to FAST in climatological comparisons. We do not compare our climatology to other TC hazard models as rigorous hazard intercomparisons are done elsewhere \cite{meiler_intercomparison_2022}.

\begin{figure}
    \centering
    \includegraphics[width=1.0\linewidth]{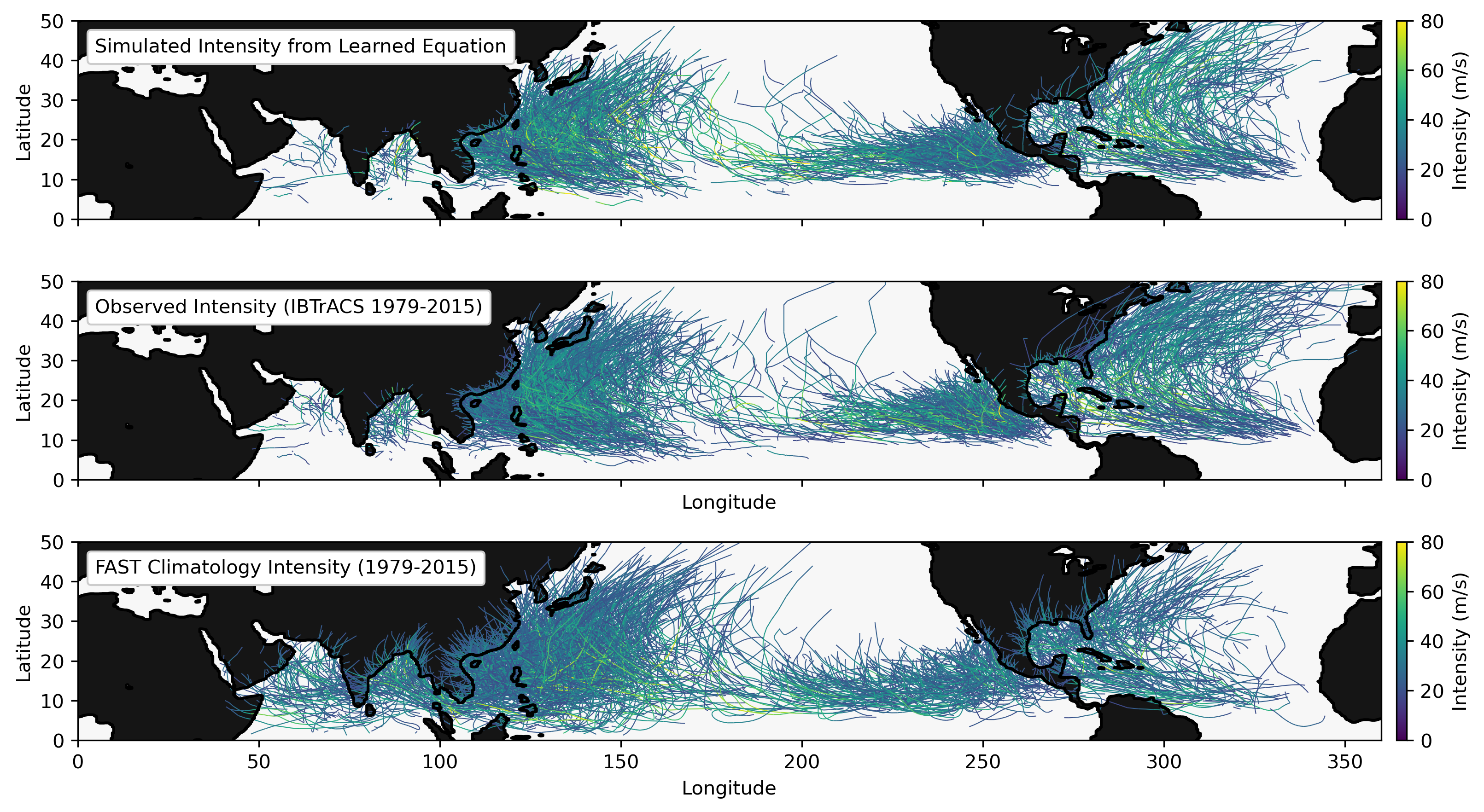}
    \caption{(Top) Synthetic intensification along a test dataset of all IBTrACS tracks from 1979-2015. (Middle) All tracks and intensities in the IBTrACS dataset, for comparison. (Bottom) Tracks generated from FAST over the same period, with the same number of total storms as in observations.}
    \label{fig:synthetic-ibtracs}
\end{figure}

In Fig. \ref{fig:synthetic-ibtracs}, we plot a single synthetic intensification series for each track in the testing dataset, restricted to times when the storm is a Tropical Storm or stronger ($\geq18\si{\meter\per\second}$). Visually, the most intense simulated storms are more intense than the most intense in IBTrACS, but the correspondence between storms at lower intensities is strong.

Next, we quantitatively assess climatological estimates of cyclone hazard. The Power Dissipation Index is a commonly used metric of hazard which is the integral of the cube of intensity over a cyclone's lifetime

\[PDI=\int_0^\tau v(t)^3dt.\]

The cube of intensity quantitatively captures that more extreme cyclones pose much greater hazard than smaller cyclones \cite{emanuel_increasing_2005}.

\begin{figure*}
    \centering
    \includegraphics[width=1.0\linewidth]{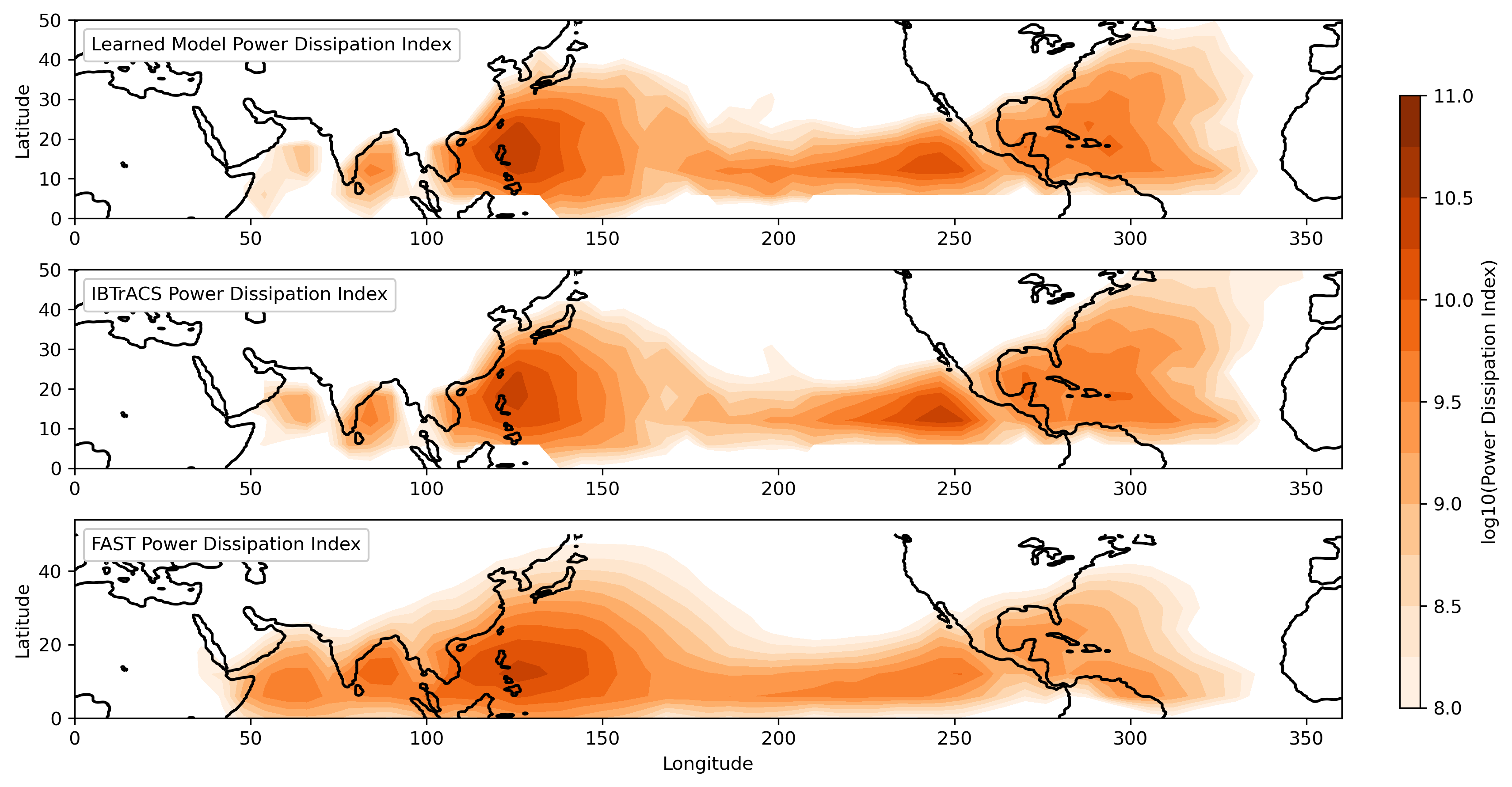}
    \caption{Annually averaged Power Dissipation Index climatology for learned model (top) and IBTrACS (middle) and FAST (bottom). PDI is computed in $6^\circ\times6^\circ$ boxes the contour plotted.}
    \label{fig:PDI-climatology}
\end{figure*}

The Power Dissipation Index climatology for the Northern hemisphere using the learned model and observations is plotted in Figure \ref{fig:PDI-climatology}. The PDI for a given year in a given basin is the sum of the integral above applied to every TC in a basin. To compute the figure, we treat each $6^\circ\times6^\circ$ degree region as a basin and take an annual average, using 1 simulated storm per IBTrACS track per year. The mean PDI over the domain for the learned climatology is $109.78\%$ of the observational climatology, while the Mean PDI for the FAST climatology is $93.33\%$ of the observed. Our model somewhat overestimates intensity while FAST somewhat underestimates it.

We also investigate the Lifetime Maximum Intensity Distribution for IBTrACS and the learned model in Fig. \ref{fig:lmi}. Agreement is strong for all storms at or below $65\si{\meter\per\second}$, but there is overestimation of extreme LMIs. This can be attributed to the standard deviation of stochastic perturbations increasing linearly with intensity as determined in Sec. \ref{sec:calibration}, which likely makes perturbations too extreme at extreme intensities and artificially inflates LMI. FAST somewhat underestimates LMI for Category 3 and above ($\geq50\si{\meter\per\second}$).

\subsection{Landfalling Statistics}

\begin{figure}
    \centering
    \includegraphics[width=1.0\linewidth]{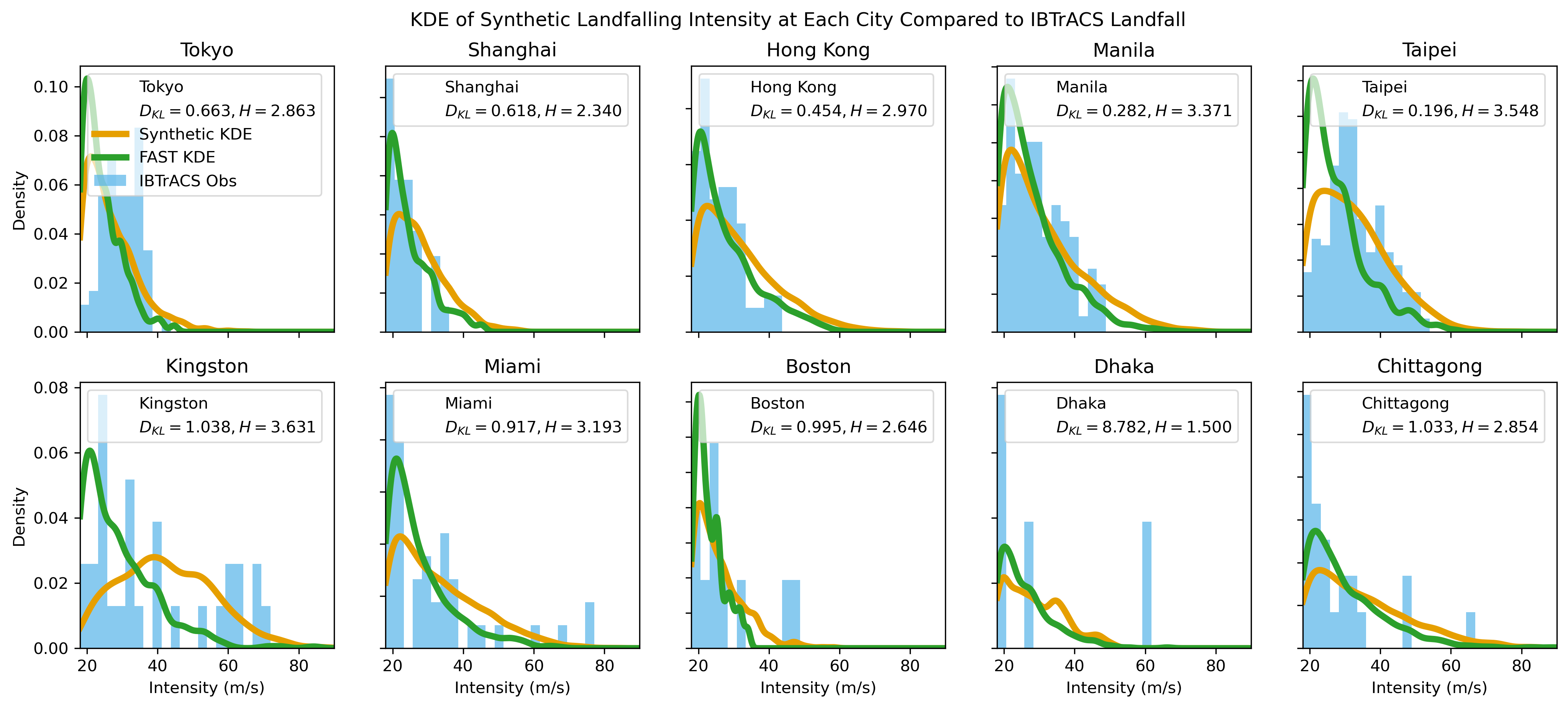}
    \caption{Intensity distribution of storms a $150\si{\kilo\meter}$ radius of a collection of major cities for IBTrACS historical data, from the learned model, and FAST. The KL Divergence from the historical to synthetic KDE is reported in each case.}
    \label{fig:landfalling-intensity}
\end{figure}

Next, we test the statistics of intensity at landfall. In Fig. \ref{fig:landfalling-intensity}, we estimate a continuous density of intensity at landfall for a collection of populous coastal cities, measured as all intensity values of storms within $150\si{\kilo\meter}$ of the city. Next, we estimate a histogram of landfalling intensities based on the whole IBTrACS dataset, using the $5$ knot discretization of IBTrACS to set bin sizes. Then, we compute the KL-Divergence of the synthetic KDE against the IBTrACS histogram

\begin{equation}
    D_{KL}(p_{obs}||q_{synth})=\sum_i p_{obs}(v_i)\log_2\left(\frac{p_{obs}(v_i)}{q_{synth}(v_i)}\right)
\end{equation}

where $p_{obs}$ is the histogram of IBTrACS intensities at landfall and $q_{synth}$ is the modelled density. We report the divergence in bits as well as the entropy $H=\sum_ip_i\log_2(1/p_i)$ of the IBTrACS histogram for comparison. We find that the match is strong for all cities except for Tokyo, where intensity is clearly underestimated, and Dhaka, where the most extreme storm observed is afforded low probability. FAST behaves similarly but affords more probability mass to weaker storms.

\subsection{Return Period Assessment}\label{sec:return-periods}

\begin{figure}
    \centering
    \includegraphics[width=1.0\linewidth]{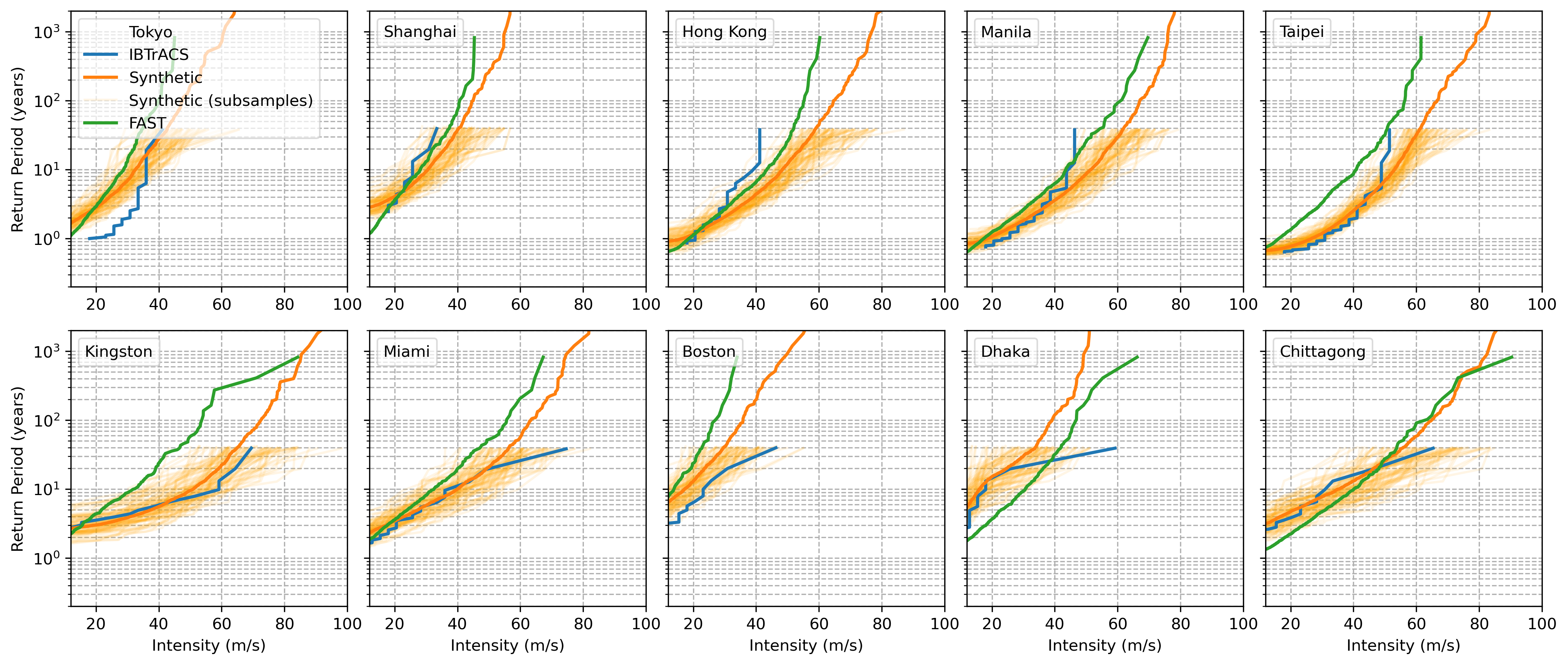}
    \caption{Return period curves for a radius of $150\si{\kilo\meter}$ region around a large collection of populous cities in the Northern Hemisphere.}
    \label{fig:return-periods}
\end{figure}

We compute return period curves for a collection of important cities in Fig. \ref{fig:return-periods}. We utilize the Weibull formula  \cite{makkonen_plotting_2006} which calculates the annual exceedance probability as 

\begin{equation}
    P(v_i)=\frac{i}{n+1}\frac{n}{m}
\end{equation}

where $v_i$ is the $i$th highest intensity in a region over the simulated climatology duration, $n$ is the total number of storms in that region, and $m$ is the duration of the climatology in years. The return period is then the reciprocal of the exceedance probability. Return periods are computed over the IBTrACS dataset from 1979 to 2015 and over the testing dataset with 100 synthetic tracks per real track, which we treat as 3600 years of synthetic storms. The maximum intensity of storms while they are within $150\si{\kilo\meter}$ of the city of interest is retained.

There is agreement between observed and synthetic return periods for Shanghai, Kingston, Miami, Boston, Dhaka and Chittagong, but disagreement in Tokyo, Hong Kong, Manila and Taipei. In Tokyo, the return period of storms under $35\si{\meter\per\second}$ is overestimated. In Hong Kong, Manila and Taipei - all Tropical cities in the North Western Pacific basin - return periods of storms above $45\si{\meter\per\second}$ are significantly overestimated compared to observations, suggesting a positive bias in the most extreme storm events in this basin, which can again be attributed to our stochastic parameterization from Sec. \ref{sec:calibration} as with the LMI distribution. Meanwhile, the return period of storms approaching Tokyo is overestimated, perhaps due to the inclusion of a damping term $V_pS$ which scales damping due to wind shear by the potential intensity which is highest in the Western Pacific basin.

\begin{figure}
    \centering
    \includegraphics[width=0.8\linewidth]{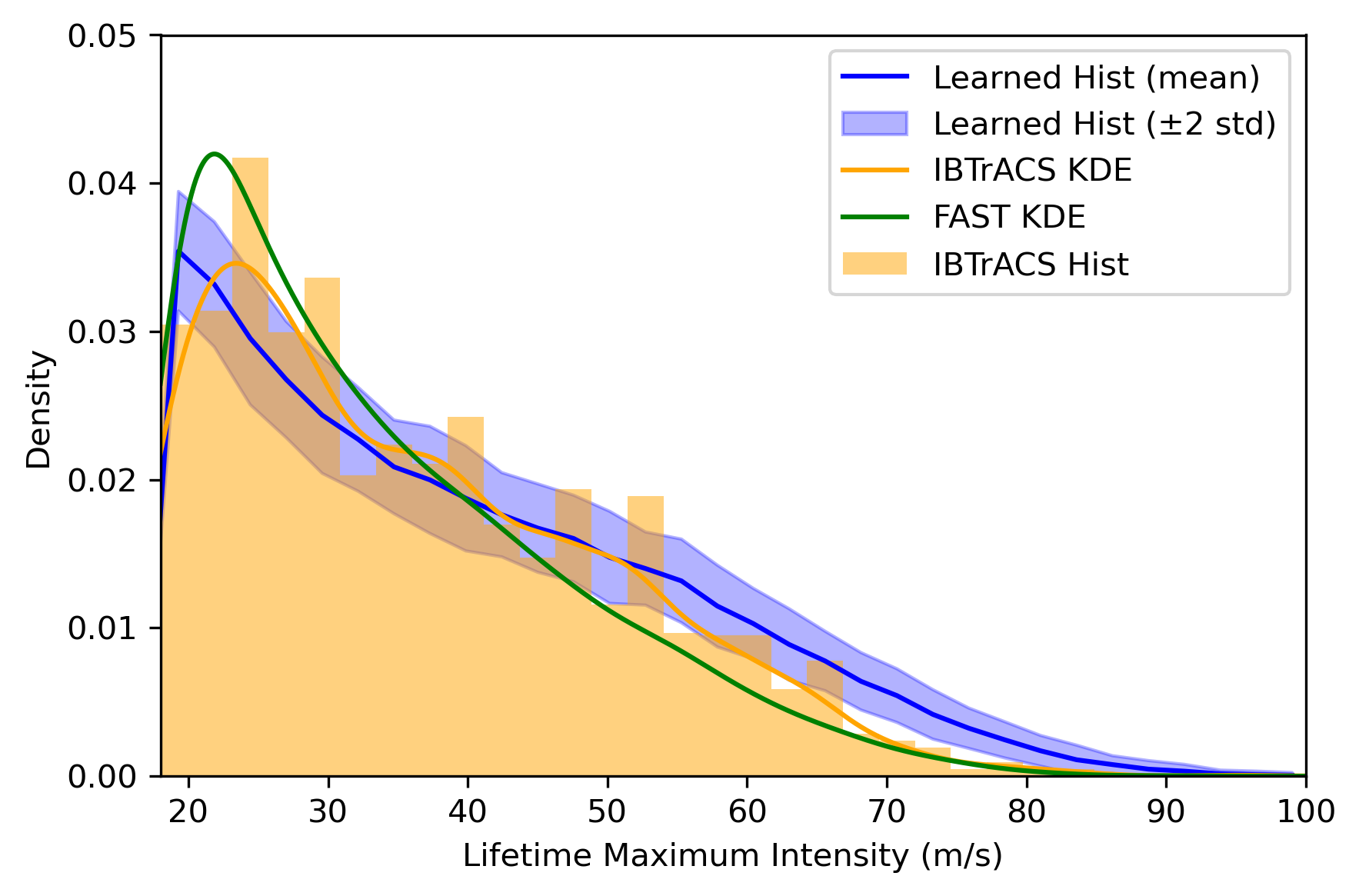}
    \caption{The Lifetime Maximum Intensity distribution of the IBTrACS dataset in the Northern Hemisphere, from 1979-2015. Corresponding distribution for synthetic climatology generated from the learned model with uncertainty found by subsampling the learned model to the size of the IBTrACS dataset. Also plotted is the LMI distribution from a FAST climatology.}
    \label{fig:lmi}
\end{figure}

\section{Equation Learning}
\label{sec:equation-learning}

Equation learning proceeds in 3 steps. First, we select features by solving an Integral SINDy problem \cite{brunton_discovering_2016, schaeffer_sparse_2017}. Second, we refit the parameters of the chosen features using an Ensemble Kalman Filter as a fine-tuning step \cite{trautner_informative_2020}. Third, we calibrate a stochastic component to represent unresolved intensification processes.

\subsection{Integral SINDy}

We use Integral SINDy \cite{schaeffer_sparse_2017} to identify the subset of terms for use in the intensification model. Integral SINDy sets up a sparse linear regression problem by using the integral formulation of an inhomogeneous differential equation $\dot{x}=F(x,t)$ which is 

\begin{equation}\label{eq:ode-integral}
    x(t_i+T)-x(t_i)=\int_0^TF(x(t'+t_i),t'+t_i)dt'
\end{equation}

for a time horizon $T$ and for any initial time $t_i$. Assuming the linear model structure for $F(x(\tau),\tau)$ from Eq. \ref{eq:model_setup}, we have that $F(x(t),t)=\sum_j\beta_j\phi_j(x(t),t)$ for features $\phi_j$ which lets us move the integral in Eq. \ref{eq:ode-integral} over all individual features

\begin{equation}
    x(t_i+T)-x(t_i)=\sum_j\beta_j\int_0^T\phi_j(x(t'+t_i),t'+t_i)dt'.
\end{equation}

If we denote each integral term as $d_{j,i,T}=\int_0^T\phi_j(x(t'+t_i),t'+t_i)dt'$ and each time difference as $C_{i,T}=x(t_i+T)-x(t_i)$, we find a set of linear equations

\begin{equation}\label{eq:integral-linear}
    C_{i,T}=\sum_j\beta_j d_{j,i,T}.
\end{equation}

We estimate the temporal differences in intensity $C_{i,T}$ directly from IBTrACS intensity time series. Each integral is approximated using piecewise constant quadrature over the 6 hour time increments in the data $\int_0^T\phi_j(x(t'+t_i),t'+t_i)dt'\approx\sum_{k=0}^{T/\Delta t}\phi_j(x(k\Delta t+t_i),k\Delta t+t_i)\Delta t$ for $\Delta t=6\si{\hour}$. Piecewise constant quadrature is used because it averages out observation noise in the integrated terms \cite{schaeffer_sparse_2017}. We use the intensity from IBTrACS and the environmental variables from ERA5 dataset to compute $\phi_j(x(k\Delta t+t_i),k\Delta t+t_i)$.

Note that a linear equation can be produced for every initial condition and for every valid time horizon $T$ from that initial condition. This means we can produce an abundance of constraints - by using all storms from 2016 to 2020 and with a maximum time horizon of $T=5\text{ days}$, we create $n=71,553$ rows. We seek a sparse solution which minimizes

\begin{equation}
    \hat{\beta}=\arg\min_\beta ||C-\beta D|| \quad \textrm{s.t.}\quad \ell_0(\beta)\leq k
\end{equation}

for some subset size $k$ and $C,D$ are flattened matrix representations of Eq. \ref{eq:integral-linear}.

This is a Best Subset Selection problem \cite{zhu_polynomial_2020}. In general, this problem is NP-Hard \cite{davis_adaptive_1997} due to concavity of the regularization penalty. A variety of greedy approximate algorithms are available such as Matching Pursuit \cite{mallat_matching_1993}, Orthogonal Matching Pursuit \cite{cai_orthogonal_2011}, and Least-Absolute Shrinkage and Selection Operator (LASSO) \cite{hastie_statistical_2015}. Advances in Mixed Integer Quadratic Programming solvers has enabled tractable global solutions for some problems \cite{bertsimas_learning_2023}. We use the recently developed Absolute Best-Subset Selection (ABESS) algorithm because of its computational efficiency and favorable convergence properties \cite{zhu_polynomial_2020}.

\begin{figure}
    \centering
    \includegraphics[width=0.8\linewidth]{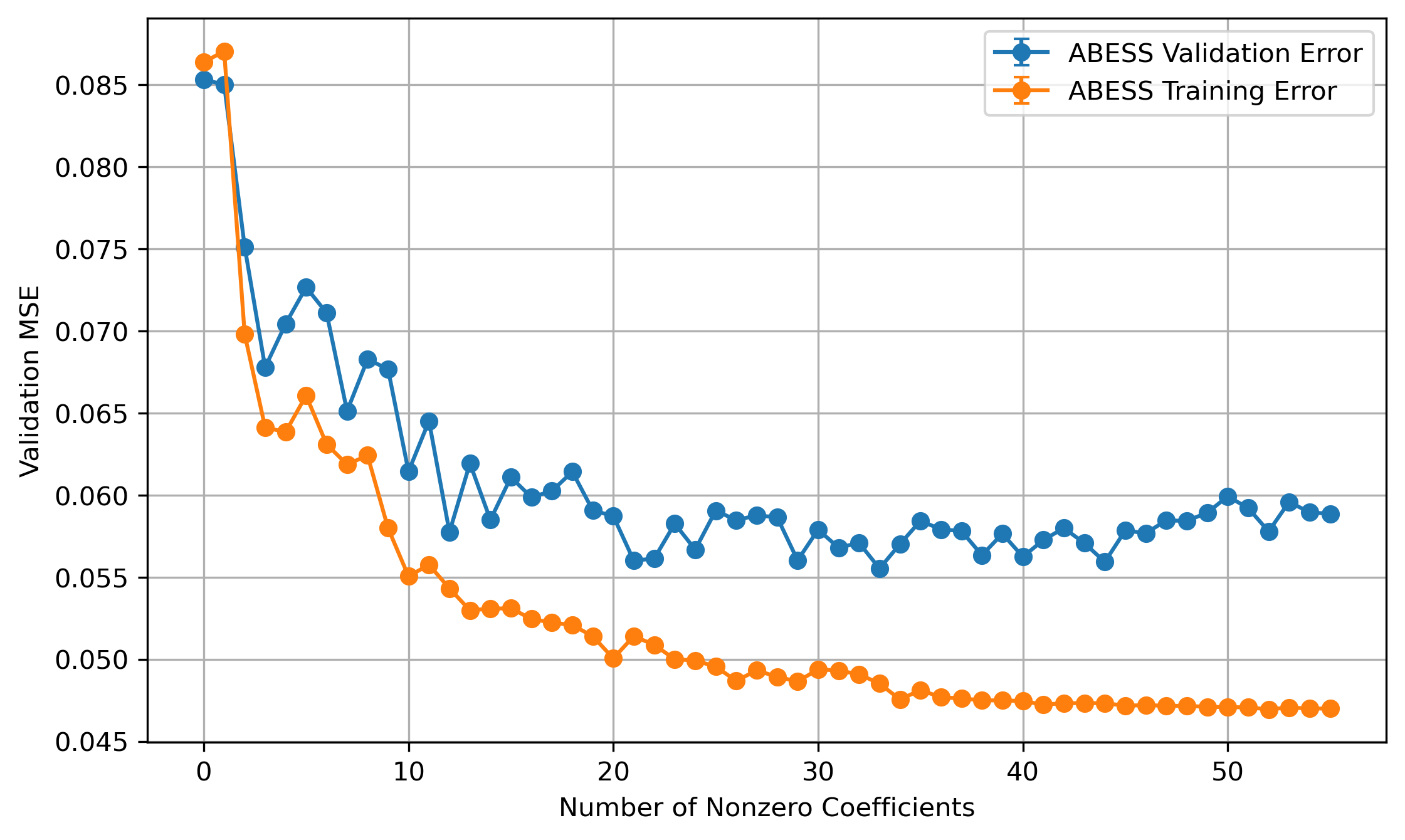}
    \caption{Validation error of sparse solutions as a function of the number of included terms on a held-out validation dataset.}
    \label{fig:sparsity_plot}
\end{figure}

To assess how many terms to include, we plot the cost on the validation dataset using a model trained on training dataset as a function of the number of included terms $k$ in Figure \ref{fig:sparsity_plot}. We select $k=10$ terms as a tradeoff between skill and model complexity and find a 38\% reduction in error compared to persistence on the validation dataset. The model is trained using IBTrACS storms from 2016 to 2020 which have been matched to ERA5 counterparts \cite{hodges_how_2017}. The training-validation split is $90\%-10\%$ and storms are chosen at random. The testing dataset is all storms from 1979 to 2015.

\subsection{Parameter Fine-Tuning}

We fine-tune the parameters for the chosen 10 terms using an Ensemble Kalman Update. In the terminology
of \cite{trautner_informative_2020}, this step realizes the ensemble-based inference component of the informative learning cycle, using the ensemble Kalman update as an adjoint-free parameter estimator for the nonlinear TC intensification model.

The Ensemble Kalman Update is a generalization of the Kalman Update which assumes that an observation is a linear function of the latent parameter of interest, contaminated by observation noise \cite{kalman_new_1960}

\begin{equation}
    y=H\beta+\nu
\end{equation}

where $H$ is a linear map and $\nu\sim\mathcal{N}(0,C_{\nu\nu})$ is Gaussian observation noise assumed independent from the underlying parameters $\beta$. The Ensemble Kalman Update extends the linear Kalman update by using an ensemble to make a linear approximation $H$ of a nonlinear observation mapping $h$.

Let the ensemble of 10 parameters $\beta$ be samples from a Gaussian prior $\beta\sim\mathcal{N}(\beta^-,C_{\beta\beta}^-)$ and denote the ensemble itself as  $X=\left[\beta_1\quad\beta_2\quad\cdots\quad\beta_E\right]$ where $E$ is our ensemble size. By integrating the intensification model defined by each $\beta_i$ from an initial condition at time $t_i$ to a later point in time $t_i+T$, which we represent by the nonlinear function $h(\cdot)$, we can compute an ensemble of predicted intensities $Y=\left[ h(\beta_1)\quad h(\beta_2)\quad\dots\quad h(\beta_E)\right]$ for an observed intensity $y=v_\text{IBTrACS}(t_i+T)$.

Given a prior ensemble over coefficients $\beta^-$ of dimension $10\times E$, the Ensemble Kalman update is

\begin{align}
    \beta^+=\beta^-+K'(y-Y)
\end{align}

where the Kalman Gain matrix $K$ \cite{roth_ensemble_2017} is computed using the ensemble as

\begin{equation}
    K'=\frac{1}{N-1}\Tilde{\beta}\Tilde{Y}^T\left(\frac{1}{N-1}\Tilde{Y}\Tilde{Y}^T+C_{\nu\nu}\right)^{-1}.
\end{equation}

We chose observation noise covariance as $C_{\nu\nu}=I\sigma^2$ where $\sigma=2.57\si{\meter/\second}$ to reflect the discretization of intensity in IBTrACS, which is binned to $2.57\si{\meter/\second}$ increments. In general, this linear approximation in the Ensemble Kalman Update is not guaranteed to converge for nonlinear systems \cite{crisan_oxford_2011}. As our observation function involves the integration of a nonlinear dynamical system, our procedure is therefore not guaranteed to converge, however, empirically the ensemble Kalman Update has been successfully used for data assimilation in many nonlinear systems in geosciences \cite{kalnay_atmospheric_2002}. After parameter fine-tuning, the model achieves a $37.6\%$ improvement in 3 day ahead intensity forecasts on the validation dataset compared to a persistence model.

\subsection{Model Training}

Our training data consists of 5 years of ERA5 data from 2016 to 2020 restricted to TCs in the Northern Hemisphere. To simulate a cyclone, the intensification ODE is forced by time-dependent environmental variables from ERA5 along the track of an IBTrACS cyclone. At a particular training step, the model's intensity is initialized to the IBTrACS intensity at a point along an observed track and the whole ensemble $[\vec{\beta}_1,\vec{\beta}_2,\dots,\vec{\beta}_E]$ is integrated 6 hours into the future. The ensemble size is $E=100$. The Kalman Learner update is batched over 16 initial conditions to mitigate the EnKF's tendency to be overconfident \cite{roth_ensemble_2017}. We then perform 11 more epochs of training, each at a time horizon from 12 hours to 3 days in 6 hour increments. Such rollout is crucial for model stability. We retain only the mean of the resulting posterior of coefficients as the final parameter values, so system stochasticity arises only from the model's stochastic component, which is learned in Sec. \ref{sec:calibration}. With CPU-only parallelization over ensemble members, this training step takes $17\si{\hour}$.


\subsection{Calibration}\label{sec:calibration}

\begin{figure}
    \centering
    \includegraphics[width=0.8\linewidth]{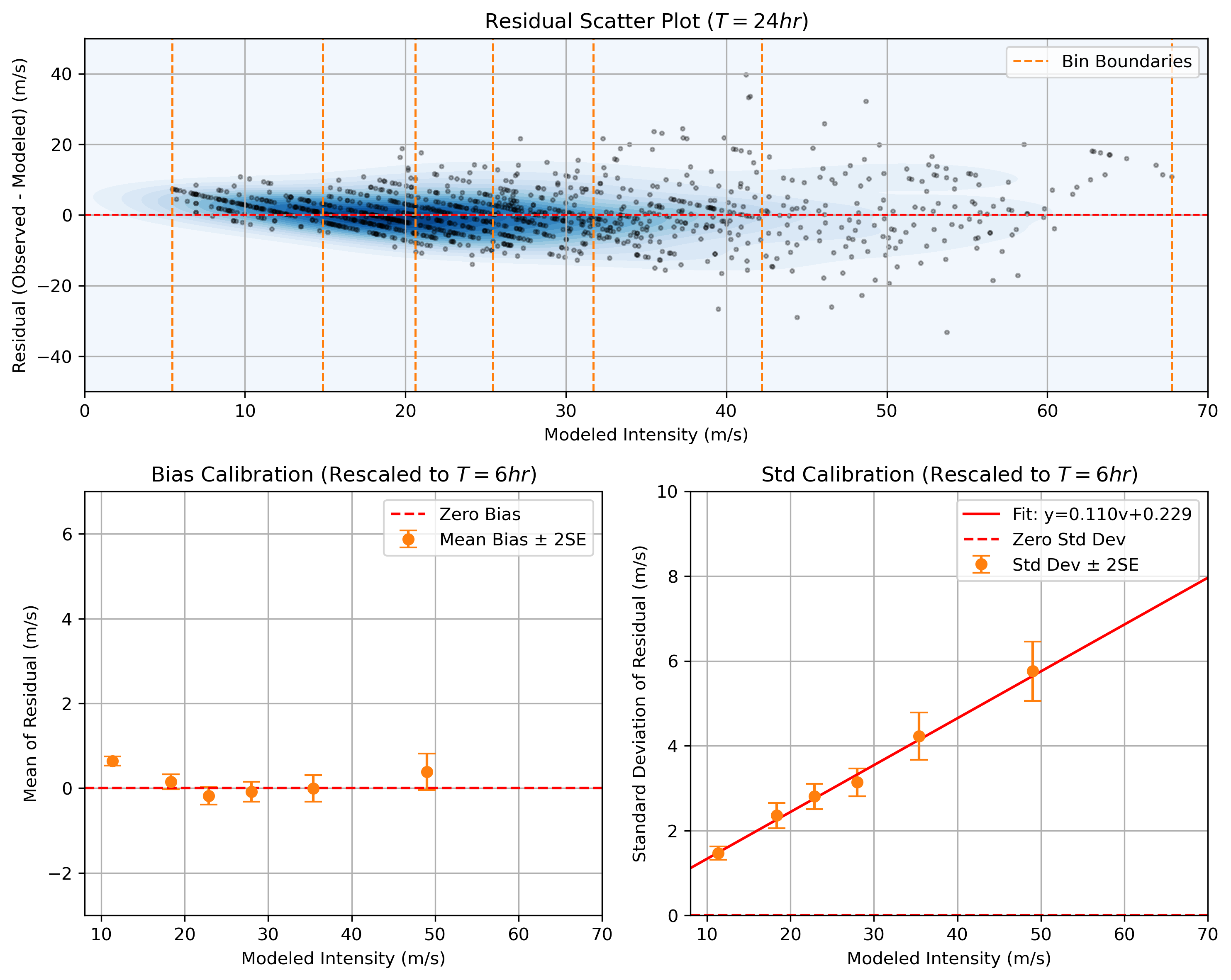}
    \caption{(Top) Distribution of residuals over a single $6\si{\hour}$ time step computed over the validation dataset. Bin boundaries are specified as orange vertical lines. (Bottom Left) The mean residual in each bin, and bootstrapped standard errors using 1000 resamplings. (Bottom Right) The residual standard deviation also with bootstrapped standard errors using 1000 resamplings. Linear fit was found using weight linear regression with weights set by the standard error bars. }
    \label{fig:calibration}
\end{figure}

Here we parameterize the unresolved components of Tropical Cyclone intensification in a calibration step. We first compute the residuals of the trained deterministic model on one day ahead forecasts on the validation dataset. A plot of the residuals is included in Fig. \ref{fig:calibration}. By first binning the residuals into 6 bins with equal numbers of samples, we observe a linear relationship between the intensity and the standard deviation of the residuals. Significant bias is present only at intensities below $14\si{\meter\per\second}$, in the range of TDs, but we do not believe this bias is genuine due to the TD reporting concerns mentioned in Sec. \ref{sec:intensification-dynamics-comparison}. Hence, we do not bias correct the model. Using linear regression, we find an expression for the standard deviation of intensity, which is $\sigma(v)=0.110v+0.229\si{\meter\per\second}$. We then nondimensionalize to $\sigma'(v')=\sigma(v)/50\si{\meter\per\second}=0.110v'+4.58\times10^{-3}$ and add the stochastic component $\sigma'(v')dW_t$ to the model.

While the empirical calibration captures stochasticity at intensities below $42\si{\meter\per\second}$ (the lower boundary of the most intense bin), we do not have enough data to assess how well it extrapolates to more extreme intensities. This choice likely contributes to overestimation of LMI (Fig. \ref{fig:lmi}) and overestimation of extreme storm return periods (Fig. \ref{fig:return-periods}) in the Western Pacific.

\section{Discussion and Future Directions}
\label{sec:discussion}

In this paper, we showed that  

\begin{enumerate}
    \item A nonlinear system identification pipeline can identify a predictive and interpretable physics-style differential equation model of TC intensification from simulations and observations.
    \item The model generates a synthetic climatology that captures many aspects of observed TC climatology.
    \item The learned equation recovers nontrivial nonlinear dynamics of TCs, namely a saddle-node bifurcation predicted in separate work by theory and high resolution numerical simulations.
\end{enumerate}

The model produces synthetic intensity tracks which can be used to generate a Tropical Cyclone climatology (Fig. \ref{fig:synthetic-ibtracs}). Over historical tracks, the synthetic climatology matches observations for much of the Lifetime Maximum Intensity distribution (Fig. \ref{fig:lmi}), return period curves for many cities (Fig. \ref{fig:return-periods}) and intensities of storms at landfall (Fig. \ref{fig:landfalling-intensity}). The model exhibits notable biases, such as overestimation of return periods for weak storms about Tokyo, underestimation of return periods of the strongest storms in the North Western Pacific basin, and overestimation of lifetime maximum intensities above $65\si{\meter\per\second}$. These biases likely come from our strong structural assumptions, such as a sparse polynomial deterministic component (Sec. \ref{sec:learned-model}) and linearity in the stochastic noise standard deviation (Sec. \ref{sec:calibration}). Though this is a machine learning model, we intentionally imposed a restricted model structure for the sake of interpretability which could be sacrificed with a more flexible structure, such as a Neural Network, to further alleviate bias. Further, greater access to information (whether from more historical data, CAM simulations or theory) is needed to verify the stochasticity parameterization at extreme intensity values. That said, our stochastic parameterization does capture the process of rapid intensification, which is a key process for representing the most extreme storms \cite{lee_rapid_2016}. Rapid intensification cannot be predicted precisely from our set of coarse-grained environmental variables \cite{mercer_atlantic_2017}, however, prior work which forecasts Rapid Intensification using memory of past TC states has shown enhanced skill \cite{yang_long_2020}, which indicates RI could be better predicted with the inclusion of latent variable tracking the state of the storm's core. Adding a learned latent variable to our intensity model, such as inner core moisture as in the FAST model \cite{emanuel_fast_2017}, could enhance the representation of RI and model skill more generally.

The model, being a polynomial SDE, is ripe for mathematical analysis. A fixed point characterization demonstrated that the model captures the expected physical dependence of the environmental parameters on TC intensification. Notably, this analysis revealed a well-established but nontrivial saddle node bifurcation as parameters governing the ventilation of the storm's core are varied \cite{emanuel_physics_2026}, demonstrating that equation discovery techniques can uncover a system's dynamical structure even from limited data. On idealized systems, successful reconstruction of dynamical behavior in a learned equation, such as topological mixing in the Lorenz63 system and generalization to unseen domains, is recovered for equation models with degrees of freedom matching those in the true system \cite{li_neural_2023}. Our successful recovery of TC dynamical behavior suggests that our sparse equation-based model's degrees of freedom match those of TC intensification itself. In contrast, we expect that overparameterized neural network models may struggle to recover dynamical behavior. Beyond existing demonstrations that equation-discovery techniques can identify predictive dynamical models from temporally coarse and noisy scientific observations \cite{fasel_ensemble-sindy_2022}, these results show that equation discovery techniques can identify interesting dynamical structure. Equation learning techniques are therefore promising investigative tools to study dynamics in scientific problems, even under severe data constraints.

The model has potential to be used in hazard estimation pipelines. It can be run very quickly as it is a concise SDE and, in principle, the model could be forced by monthly-mean or scenario-based climatological fields, making it lightweight for hazard projections once suitable environmental estimates are available. Separate genesis and track generation components will need to be specified to complete this pipeline, which can also be learned with system identification approaches in future work. However,  tests of its extrapolation ability (say under climate-change change forcings or in Southern Hemisphere basins) and closer scrutiny of its regional biases are needed before it can be safely used in a hazard pipeline.

Though a performant model has been learned, it depends on pre-engineered features which greatly simplified the model learning process. Application of system identification techniques to less well understood problems, such as Extratropical Cyclone intensification, will require an additional feature learning step from primitive climate model outputs. Such a pipeline has the potential to identify novel physical features which govern intensification and present them in an interpretable low-order model which could enhance predictive skill and physical understanding.

We hope that this work opens the door to using nonlinear system identification techniques to identify data and theory driven models of Earth System dynamics. Such models may find otherwise opaque predictive regularities, represent them in a transparent form, and hence accelerate physical model development.

\subsection{Data and Code Availability}

All the code for this project is available from Github at \url{https://github.mit.edu/essg/DOLMa}, reach out to the authors to request access. The ERA5 dataset and IBTrACS datasets can be downloaded online. The ERA5 downloading scripts are from the open source Tropical Cyclone Risk project \url{https://github.com/linjonathan/tropical_cyclone_risk.git} \cite{lin_open-source_2023}.

\subsection{Acknowledgements}

This research is part of the MIT Climate Grand Challenges Jameel Observatory CREWSNet and Weather and Climate Extremes projects. Schmidt Sciences, LLC, and the Bangkok Bank generously supported this project. 

We thank Kerry Emanuel, Jiangchao Qiu, Anamitra Saha, Raffaele Ferrari, Abigail Bodner, and Talia Tamarin-Brodsky for comments and suggestions which improved the manuscript. The authors used Grammarly and OpenAI ChatGPT as a writing assistant for language polishing and to suggest edits to exposition. All technical content, claims, and errors remain the responsibility of the authors. 

\subsection{Conflict of Interest Statement}

The authors declare no conflict of interest.

\printbibliography

\end{document}